\documentclass[10pt,twocolumn,letterpaper]{article}

\usepackage[pagenumbers]{cvpr} 

\usepackage{graphicx}
\usepackage{amsmath}
\usepackage{amssymb}
\usepackage{booktabs}
\usepackage{array}

\usepackage{cuted}
\usepackage{capt-of}
\usepackage{dsfont}

\newcommand{\PreserveBackslash}[1]{\let\temp=\\#1\let\\=\temp}

\newcolumntype{C}[1]{>{\PreserveBackslash\centering}p{#1}}
\newcolumntype{R}[1]{>{\PreserveBackslash\raggedleft}p{#1}}
\newcolumntype{L}[1]{>{\PreserveBackslash\raggedright}p{#1}}

\DeclareMathOperator*{\argmin}{arg\,min}

\usepackage[pagebackref,breaklinks,colorlinks]{hyperref}

\usepackage[capitalize]{cleveref}
\crefname{section}{Sec.}{Secs.}
\Crefname{section}{Section}{Sections}
\Crefname{table}{Table}{Tables}
\crefname{table}{Tab.}{Tabs.}

\begin{document}

\title{Text and Image Guided 3D Avatar Generation and Manipulation}

\author{Zehranaz Canfes\thanks{Equal contribution.} \quad M. Furkan Atasoy\footnotemark[1] \quad Alara Dirik\footnotemark[1] \quad Pinar Yanardag\\
Boğaziçi University\\
Istanbul, Turkey\\
{\tt\small \{zehranaz.canfes, muhammed.atasoy, alara.dirik\}@boun.edu.tr, yanardag.pinar@gmail.com}
}

\maketitle


\vspace*{-\baselineskip}
\begin{strip} 
  \hspace{0.6cm} Original \hspace{1.5cm} Old \hspace{1.6cm} Young  \hspace{1.3cm} Makeup \hspace{1.2cm} Happy \hspace{1.1cm} Surprised \hspace{1.0cm} Beyonce \hspace{1.3cm}
  \begin{center}
    \includegraphics[width=2\columnwidth]{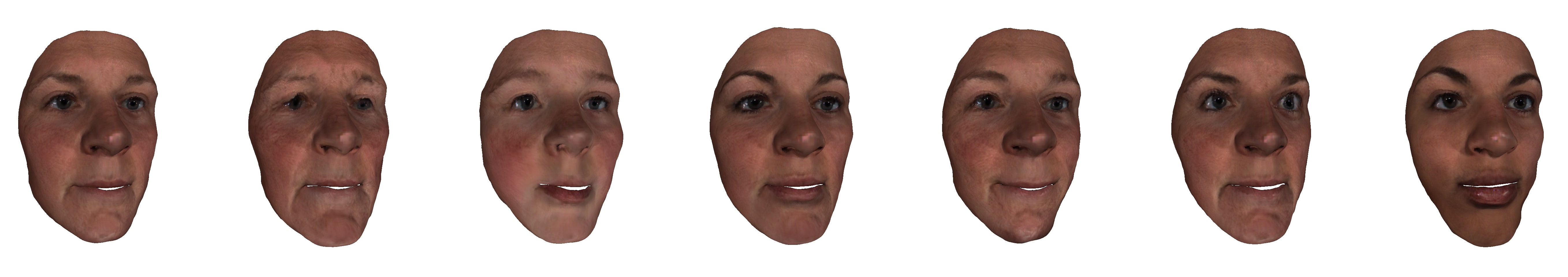}
    \captionof{figure}{Given an input vector of a 3D mesh (denoted as \textit{Original}) our method modifies the latent code such that the target attribute specified by a text prompt such as \textit{`Young'} or \textit{`Surprised'} is present or enhanced, while leaving other attributes largely unaffected.}
\label{fig:teaser}
\end{center}
\end{strip}

\begin{abstract}
The manipulation of latent space has recently become an interesting topic in the field of generative models. Recent research shows that latent directions can be used to manipulate images towards certain attributes. However, controlling the generation process of 3D generative models remains a challenge. In this work, we propose a novel 3D manipulation method that can manipulate both the shape and texture of the model using text or image-based prompts such as \textit{'a young face'} or \textit{'a surprised face'}. We leverage the power of  Contrastive Language-Image Pre-training (CLIP) model and a pre-trained 3D GAN model designed to generate face avatars, and create a fully differentiable rendering pipeline to manipulate meshes. More specifically, our method takes an input latent code and modifies it such that the target attribute specified by a text or image prompt is present or enhanced, while leaving other attributes largely unaffected. Our method requires only 5 minutes per manipulation, and we demonstrate the effectiveness of our approach with extensive results and comparisons. Our project page and source code can be found at \url{https://catlab-team.github.io/latent3D}.
\end{abstract}

\section{Introduction}
Generative Adversarial Networks (GAN) for 2D vision have achieved several breakthroughs, such as Progressive GAN \cite{karras2018progressive}, BigGAN \cite{BigGAN}, and StyleGAN \cite{Karras2019ASG, Karras2020AnalyzingAI}, which enable high-resolution and high-quality image generation in various domains. 3D vision and the field of 3D generation have made similarly remarkable progress, with the development of implicit surface and volume representations \cite{Park2019DeepSDFLC,Chen2019LearningIF,Mescheder2019OccupancyNL} enabling the encoding, reconstruction, and generation of detailed models of watertight surfaces without suffering from the limitations of using a 3D grid or fixed-topology meshes. While these implicit representation-based approaches result in a learnable surface parameterization that is not limited in resolution, they often require coordinate sampling for non-differentiable point cloud and mesh generation, which is also time consuming. Other works such as UV-GAN \cite{Deng2018UVGANAF}, GANFit \cite{Gecer2019GANFITGA}, and TBGAN \cite{gecer2020synthesizing} restrict the 3D generation problem to a 2D domain and aim to generate 3D shapes by training GANs directly on UV maps of shapes, normals, and textures.
Despite advances in 3D generation, the question of how to control the results of 3D generative models remains an active research topic. The issue of 3D manipulation is particularly important in morphable models such as the human face and body, where a natural consequence of this work is to enable animation. Previous work, known as 3D morphable models (3DMM) \cite{Blanz1999AMM, Brunton2014ReviewOS}, represent 3D faces as disentangled PCA models of geometry, expression, and texture, and manipulates faces by editing each modality separately. However, the linear nature of PCA makes it difficult to generate novel faces and high-quality reconstructions. Moreover, numerous previous work use 3DMMs as backbone models and attempt to reconstruct 3D faces from 2D images or partial face scans \cite{Wu2008FaceAV, Jourabloo2015PoseInvariant3F, Jourabloo2016LargePoseFA, Jourabloo2017PoseInvariantFA} and thus suffer from their fundamental limitations. In contrast, there have been significant advances in the manipulation of rigid 3D objects in recent years. Several methods have been proposed for manipulating implicit 3D shape representations with text \cite{Michel2021Text2MeshTN, Wang2021CLIPNeRFTD, Sanghi2021CLIPForgeTZ} and sketches \cite{Guillard2021Sketch2MeshRA}. However, these methods require hours of optimization per manipulation and are limited to rigid and simple shapes such as \textit{chairs}, while we attempt to manipulate articulated, diverse, and complex shapes such as the \textit{human face}.

In this work, we propose a method for fast and highly effective text and image-driven 3D manipulation of facial avatars. Our method uses a pre-trained generative 3D model, TBGAN, as a base GAN model and leverages the joint image-text representation capabilities of Contrastive Language-Image Pre-training (CLIP) \cite{radford2021learning} to optimize a latent code based on a user-provided text or image prompt (see Figure \ref{fig:teaser} for example manipulations). Unlike previous work \cite{Michel2021Text2MeshTN, Wang2021CLIPNeRFTD, Sanghi2021CLIPForgeTZ}, which require a large amount of time, our method requires only 5 minutes per manipulation and enables text- and image-driven edits that allow precise, fine-grained, and complex manipulations such as modifying the \textit{gender} and \textit{age} attributes without affecting the irrelevant attributes or identity of the original mesh. Our proposed method directly optimizes the shape, normal, and texture images and performs disentangled manipulations. Furthermore, we propose a baseline method that uses PCA to detect unsupervised latent directions on facial avatars. Our experiments show that our method is able to outperform PCA-based baseline and TBGAN on various simple and complex manipulations.

\section{Related Work}
\label{sec:related}

\subsection{3D Shape Representations}
Unlike 2D vision problems, where RGB images have become almost the standard data format, how to best represent 3D data remains an active research question. As a result, a variety of representations are used in work on 3D vision problems, such as point clouds, voxels, meshes, and more recently, neural implicit representations. 

One of the most popular 3D data formats is point clouds, which are lightweight 3D representations consisting of coordinate values in (x, y, z) format. They are widely used in 3D learning problems such as 3D shape reconstruction \cite{fan2017point,qi2017pointnet,Shi2019PointRCNN3O,Lin2018LearningEP}, 3D object classification \cite{fan2017point, qi2017pointnet}, and segmentation \cite{qi2017pointnet}. However, point clouds provide limited information about how points are connected and pose view- dependency issues. Another 3D format, triangular mesh, describes each shape as a set of triangular faces and connected vertices. Meshes, while better suited to describing the topology of objects, often require advanced preprocessing steps to ensure that all input data has the same number of vertices and faces. The voxel format describes objects as a volume occupancy matrix, where the size of the matrix is fixed. While the voxel format is well suited for CNN-based approaches, it requires high resolution to describe fine-grained details. Finally, numerous neural implicit representations have been proposed in recent years to overcome the shortcomings of classical representations. These methods represent 3D shapes as deep networks that map 3D coordinates to a signed distance function (SDF) \cite{Park2019DeepSDFLC} or occupancy fields \cite{Chen2019LearningIF,Mescheder2019OccupancyNL} to create a lightweight, continuous shape representation. However, a major drawback of implicit representations is that they require aggressive sampling and querying of 3D coordinates to construct surfaces. Finally, works such as UV-GAN \cite{Deng2018UVGANAF}, GANFit \cite{Gecer2019GANFITGA}, and TBGAN \cite{gecer2020synthesizing} represent shapes and textures as 2D positional maps that can be projected back into 3D space, and leverage recent advances in 2D imaging \cite{karras2018progressive} to jointly generate novel face shapes and textures. In our work, we use TBGAN as our base generative model due to its speed and generative capabilities.

\subsection {Latent Space Manipulation}
Latent space manipulation methods for image editing can be categorized into supervised and unsupervised methods. Supervised approaches typically leverage pre-trained attribute classifiers or train new classifiers using labeled data to optimize a latent vector and enhance the target attribute's presence in the generated image \cite{goetschalckx2019ganalyze,shen2020interfacegan}. On the other hand, several unsupervised approaches have been proposed to show that it is possible to find meaningful directions in the  latent space of large-scale GANs without using classifiers or labeled data \cite{voynov2020unsupervised,jahanian2019steerability}. For instance, GANSpace \cite{harkonen2020ganspace} proposes applying principal component analysis (PCA) \cite{wold1987principal} on a set of randomly sampled latent vectors extracted from the intermediate layers of BigGAN and StyleGAN. SeFA \cite{shen2020closed} proposes a similar approach that directly optimizes the intermediate weight matrix of the GAN model in closed form. A more recent work, LatentCLR \cite{yuksel2021latentclr}, proposes a contrastive learning approach to find unsupervised directions that are transferable to different classes.  Moreover, StyleCLIP \cite{patashnik2021styleclip} and StyleMC \cite{Kocasari2021StyleMCMB} both propose using CLIP for text-guided manipulation of both randomly generated and encoded images with StyleGAN2. These methods show that it is possible to use CLIP for fine-grained and disentangled manipulations of images.

\subsection{3D Shape Generation and Manipulation}
\label{sec:gan}
In recent years, there have been tremendous advances in 3D shape generation. While some of this work includes traditional 3D representations such as point clouds \cite{fan2017point, qi2017pointnet, Achlioptas2018LearningRA, Hui2020ProgressivePC, Shu20193DPC}, voxels \cite{kar2015category, choy20163d}, and meshes \cite{han2016mesh, groueix2018,MeshCNN2019}, several approaches have been proposed to use implicit surface and volume representations for high-quality and scalable representations (see \cite{Park2019DeepSDFLC, Chen2019LearningIF, Mescheder2019OccupancyNL}). However, most of this work focuses on generating rigid objects, a relatively simple task compared to the generation of articulate, morphable shapes such as the human face and body. In contrast to rigid object generation, most work on human face generation and reconstruction uses linear statistical models known as 3DMMs. \cite{Blanz1999AMM} which trains separate linear statistical models using PCA for face shape, expression, and texture on a dataset of registered face meshes where the corresponding keypoints are available. However, the linear nature of PCA makes it difficult to perform high-quality reconstructions and novel manipulations. Several works on face generation address this problem and propose various methods with limited success (see \cite{Booth20173DFM,Tran2017RegressingRA,Duong2015BeyondPC,Tewari2018SelfSupervisedMF}). In addition, 3DMM is widely used as the backbone of various applications such as 3D reconstruction of faces from multiple images \cite{Roth2016Adaptive3F,Tewari2019FMLFM,Bai2020DeepFN,Ramon2021H3DNetFH} or a single image \cite{Sela2017UnrestrictedFG,Jiang20183DFR,Xing2019ASB}.

Despite the advances in 3D generation, the work on 3D shape manipulation is much more limited and focuses on supervised or unsupervised manipulation of shapes or textures separately. Unsupervised 3D manipulation methods often introduce additional constraints into the training process to enforce better controllability \cite{Li2021SPGANS3,Elsner2021IntuitiveSE}. In addition, several supervised 3D manipulation methods have recently been proposed. Text2Mesh \cite{Michel2021Text2MeshTN} proposes a neural style model that encodes the style of a single mesh and uses a CLIP -based method for texture manipulations. However, this method requires training a separate model for each shape to be manipulated and is limited to texture manipulations. Another work, \cite{Wang2021CLIPNeRFTD}, proposes a CLIP -based method for text- and image-based manipulation of NeRFs. However, this method requires training multiple models per text to map the CLIP embedding of an input image or text onto the latent space of the proposed deformation network. Similarly, CLIP -Forge \cite{Sanghi2021CLIPForgeTZ} trains an auto-encoding occupancy network and a normalizing flow model to connect the CLIP encodings of 2D renders of simple 3D shapes such as \textit{chair} or \textit{table} and the latent shape encodings. We note that this method is limited to shape manipulation of simple rigid objects and does not allow for high-resolution generation or fine-grained edits due to the use of the voxel format. In addition, this method requires more than 2 hours per manipulation. Unlike other CLIP -based 3D manipulation methods, our method can manipulate both shape and texture and requires only 5 minutes to perform complex and accurate manipulations of articulated face avatars.

\section{Methodology}
In this section, we first briefly describe TBGAN and then introduce our method for text and image-driven manipulation of 3D objects.

\subsection{Background on TBGAN}
In our work, we use TBGAN as a generative base model for manipulation, a GAN model with an architecture that closely resembles PG-GAN \cite{Karras2018ProgressiveGO}. More specifically, TBGAN proposes a progressively growing architecture that takes a one-hot- encoded facial expression vector $\mathbf{e}$ encoding 7 universal expressions \textit{neutral, happy, angry, sad, afraid, disgusted, and surprised} and a random noise vector $\mathbf{z}$ as input. It then progressively generates higher-dimensional intermediate layer vectors known as \textit{modality correlation} layers, and branches to \textit{modality-specific} layers to jointly generate high-quality shape, shape-normal, and texture images. The model is trained on large-scale, high-resolution UV maps of preprocessed meshes with WGAN-GP loss \cite{Gulrajani2017ImprovedTO}. Within the modality correlation layers, the so-called \textit{trunk network} preserves the modality correspondences, while the separate branches of the modality-specific layers allow learning independent distributions for shape and texture data. Unlike 3DMMs, which are often constructed by applying PCA to datasets of 3D scans from hundreds or thousands of subjects \cite{Blanz1999AMM}, TBGAN is not bound by linear separability constraints and provides a continuous latent space.

\begin{figure*}[t!]
\begin{center}
\centerline{\includegraphics[width=1.6\columnwidth]{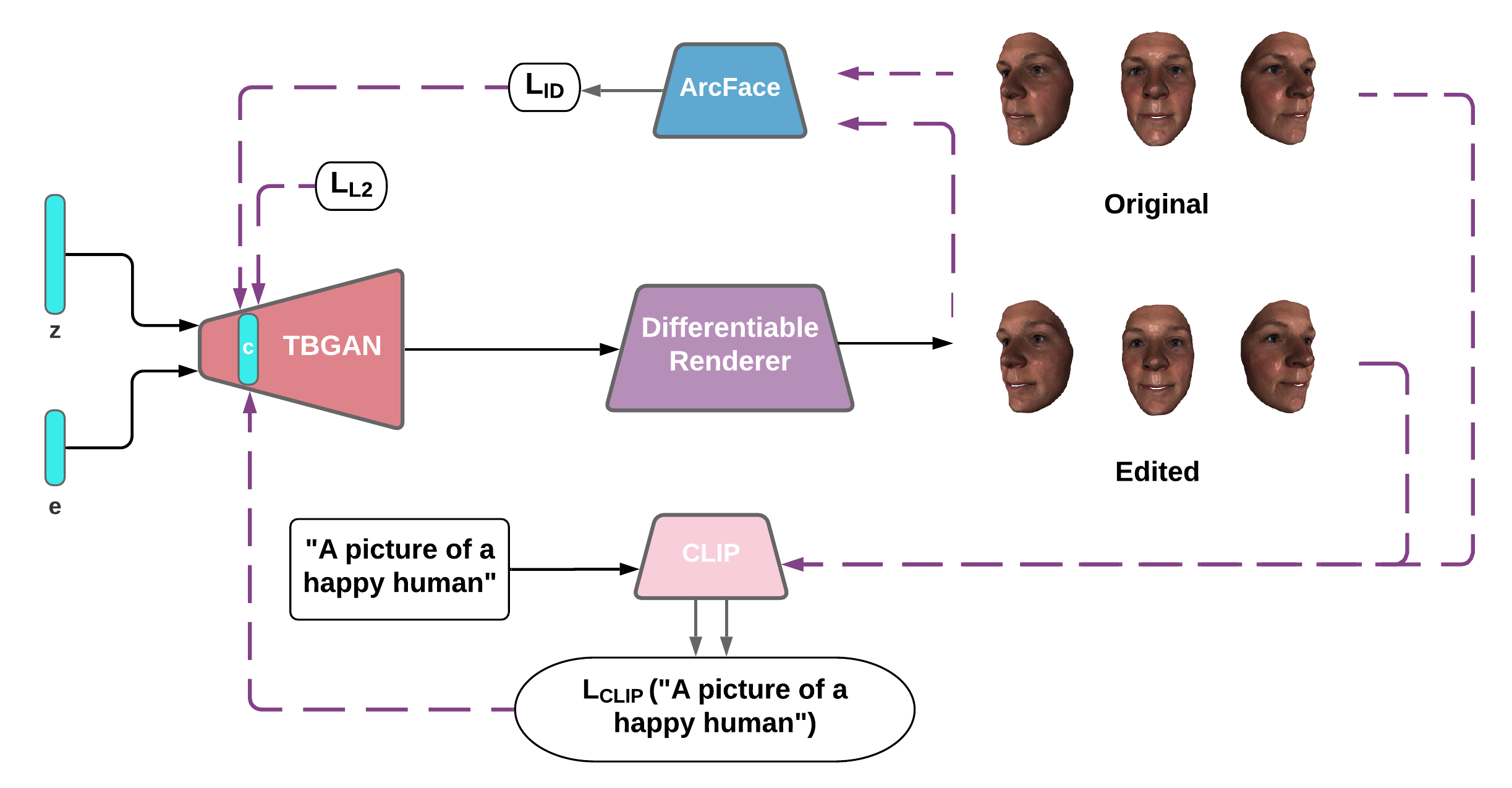}}
\caption{An overview of our framework (using the text prompt \textit{`happy human'} as an example). The manipulation direction $\Delta \mathbf{c}$ corresponding to the text prompt is optimized by minimizing the CLIP-based loss $\mathcal{L}_{\text{CLIP}}$, the identity loss $\mathcal{L}_{\text{ID}}$, and the L2 loss $\mathcal{L}_{\text{L2}}$.}
\label{fig:framework}
\end{center}
\vskip -0.3in
\end{figure*}

\subsection{Text and Image Guided Manipulation}
\label{sec:manipulation}
Given a pre-trained TBGAN generator $\mathcal{G}$, let $\mathbf{z} \in \mathcal{R}^d$ denote a d-dimensional random input vector sampled from a Gaussian distribution $\mathcal{N}\left(0, \sigma^{2}\right)$ and $\mathbf{e}$ denote a one-hot- encoded facial expression vector initialized to zero to obtain a neutral expression. Let $\mathbf{c} \in \mathcal{C}$ denote an intermediate layer vector obtained by partial forward propagation of $\mathbf{z}$ and $\mathbf{e}$ through the generator $\mathcal{G}$. Our method first generates a textured mesh by using the generated shape, normal and texture UV maps via cylindrical projection. Then given a text prompt $t$ such as '\textit{happy human}', $\mathbf{c}$ is optimized via gradient descent to find a direction $\Delta \mathbf{c}$, where $\mathcal{G}(\mathbf{c} + \Delta \mathbf{c})$ produces a manipulated textured mesh in which the target attribute specified by $t$ is present or enhanced, while other attributes remain largely unaffected.
 In our work, we optimize the original intermediate latent vector $\mathbf{c}$ using gradient descent and work in the $4\times4$/dense layer of the TBGAN generator (see ablation study in Section \ref{app:layer} on the choice of layer used for manipulation). The optimized latent vector $\mathbf{c} + \Delta \mathbf{c}$ can then be fed into TBGAN to generate shape, normal, and texture UV maps, and finally a manipulated mesh with the target attributes. A diagram of our method is shown in Figure \ref{fig:framework}. To perform meaningful manipulation of meshes without creating artifacts or changing irrelevant attributes, we use a combination of a CLIP-based loss, an identity loss, and an L2 loss as follows:

\begin{equation}
    \argmin_{\Delta \mathbf{c} \in \mathcal{C}} \mathcal{L}_{\text{CLIP}} + \lambda_{\text{ID}} \mathcal{L}_{\text{ID}} + \lambda_{\text{L2}} \mathcal{L}_{\text{L2}}
 \label{eq:final}
\end{equation}

where $\lambda_{\text{ ID }}$ and $\lambda_{\text{L2}}$ are the hyperparameters of $\mathcal{L}_{\text{ ID }}$ and $\mathcal{L}_{\text{L2}}$, respectively. While CLIP-based loss ensures that the user-specified attribute is present or enhanced, ID-loss and L2-loss leave other attributes unchanged, forcing disentangled changes. The identity loss $\mathcal{L}_{\text{ ID }}$ minimizes the distance between the identity of the original renders and the manipulated renders:

\begin{equation}
\mathcal{L}_{\text{ID}}  = 1- \left \langle R(\mathcal{G}(\mathbf{c})), R(\mathcal{G}(\mathbf{c} + \Delta \mathbf{c})) \right \rangle 
\end{equation}

where $R$ is ArcFace \cite{Deng2019ArcFaceAA}, a facial recognition network in the case of face recognition, and $\langle \cdot, \cdot \rangle$ computes the cosine similarity between the identities of the rendered image and the manipulated result (see ablation study in Section \ref{app:identity} on the effect of identity loss). The L2 loss is used to prevent artifact generation and  defined as:
\vspace{-0.15em}
\begin{equation}
\mathcal{L}_{\text{L2}}=||\mathbf{c} - (\mathbf{c} + \Delta \mathbf{c})||_2
\end{equation}

The CLIP -based loss term $\mathcal{L}_{\text{ CLIP }}$ can be defined in two different ways, depending on the type of prompt provided by the user: the user can either provide text prompts such as \textit{'old human'} or a target image such as an image of \textit{Bill Clinton} for manipulation. If the user provides a list of text prompts, the CLIP-based loss is given by:
\begin{equation}
 \mathcal{L}_{\text{CLIP}} = \frac{\sum_{j=1}^K \sum_{i=1}^N D_{\text{CLIP}}(\mathcal{I}_i, t_j)}{K \cdot N}
\end{equation}

where $\mathcal{I}_i$ is a rendered image from a list of $N$ rendered images, $t_j$ is the target text $t$ embedded in a text template from a list of $K$ templates. Here, $\mathcal{L}_{ CLIP }$ is used to minimize the cosine distance between CLIP embeddings of the rendered images $I_i$ and the set of text prompts $t_j$. In the case where the user specifies a target image, the CLIP-based loss is given by:
\begin{equation}
 \mathcal{L}_{\text{CLIP}} = \frac{\sum_{i=1}^N D_{\text{CLIP}}(\mathcal{I}_i, \mathcal{I}_{\text{targ}})}{N}
\end{equation}

\begin{figure*}
{\hspace{1.7cm} Original \hspace{1.5cm} Old \hspace{1.7cm} Child  \hspace{1.3cm} Big Eyes  \hspace{0.5cm} Thick Eyebrows  \hspace{0.7cm} Makeup}
\begin{center}
\includegraphics[width=1.8\columnwidth]{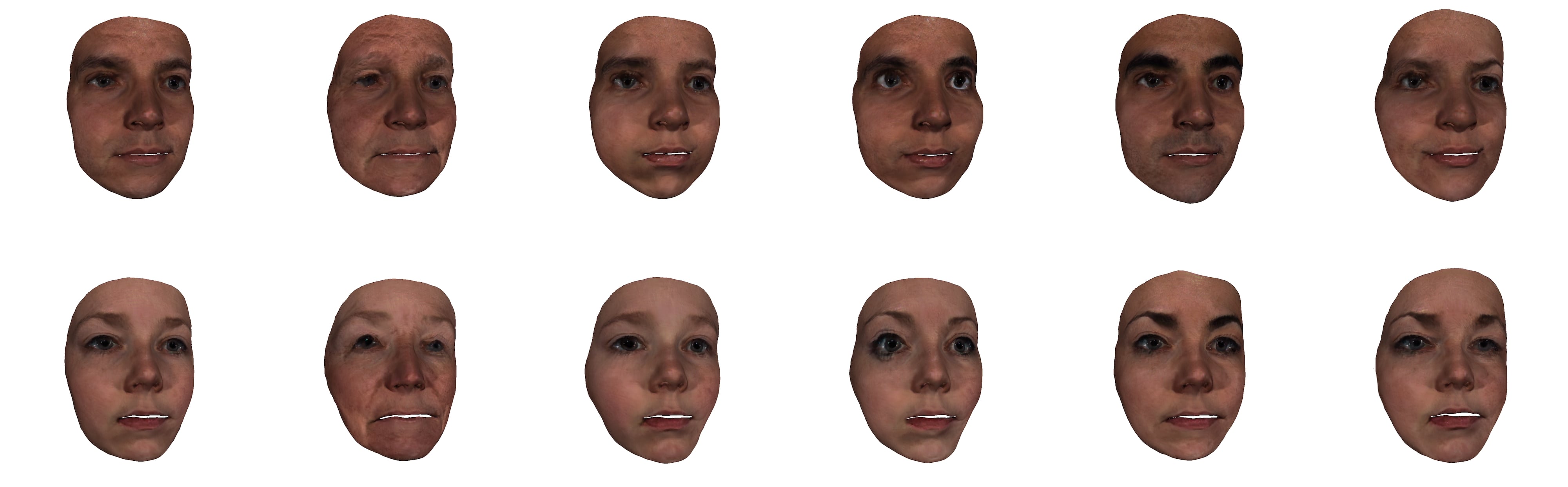}
\caption{Manipulation results of our method with various inputs and text prompts on two different 3D faces: 'Old', 'Child', 'Big Eyes', 'Thick Eyebrows', 'Makeup'. The leftmost column shows the original outputs, the adjacent columns show the manipulation results, the target text prompt is above each column.}
\label{fig:simple}
\end{center}
\vskip -0.1in
\end{figure*}
where $\mathcal{I}_{\text{targ}}$ is the target image. Here, $\mathcal{L}_{\text{ CLIP }}$ seeks to minimize the cosine distance between CLIP embeddings of the rendered images $\mathcal{I}_i$ and the target image $\mathcal{I}_{\text{targ}}$. We use $D_{\text{ CLIP } }$ to compute the cosine distance between CLIP embeddings in both methods. The renderings $\mathcal{I}_i$ and templates $t_j$ are created as follows.

\paragraph{Differentiable rendering} Note that to optimize the latent code $\mathbf{c}$, we need to compute the CLIP distance between the given text prompt and the generated mesh. Since the CLIP model cannot handle 3D meshes, we render the generated mesh corresponding to the latent code $\mathbf{c}$ in 2D and feed it with the pre-trained CLIP model. However, this simple strategy is not sufficient to optimize the latent code $\mathbf{c}$ via gradient descent, since the rendering operation needs to be differentiable. To this end, we create a fully differentiable pipeline using an open-source differentiable renderer library, PyTorch3D \cite{Ravi2020Accelerating3D}. To enforce view consistency, we render N=3 views of the generated mesh from an anchor, where the mesh $\mathcal{M}$ is rotated by $-30$, $3$, and $30$ degrees: $f_{\text{diff}}(\mathcal{M}, \theta_{cam}) = \mathcal{I}$, where $\theta_{cam}$ denotes the camera, object position, and rotation parameters used by the differentiable renderer. We denote the renders generated by $f_{\text{diff}}$ as $\mathcal{I}_1$, $\mathcal{I}_2$, and $\mathcal{I}_3$. We compute the CLIP-based loss for each image and average them prior to feeding them into our loss, which leads to more stable results.

\paragraph{Prompt engineering} 
Our model takes a user-defined text prompt $t$ as input which  describes the target manipulation. Previous work has shown that using prompt engineering \cite{radford2021learning} to augment the text prompts yield more consistent results. Hence, we augment the original text prompt $t$ by embedding it in sentence templates such as '\textit{a rendering of a \ldots}' or '\textit{a face of a \ldots}' to generate $K$ text prompts $t_1, t_2, ..., t_k$ (see Appendix \ref{app:templates} for the full list of templates used in this work). Note that more than one semantically equivalent text prompt can be given as input to achieve more stable results. For example, to achieve a manipulation that makes the face look \textit{older}, our method can use a list of different text prompts such as '\textit{old human}' and '\textit{aged person}'.

\begin{figure}
\hspace{0.4cm} Asian \hspace{1.1cm} Indian \hspace{0.9cm} Woman  \hspace{1.0cm} Man
\begin{center}
\centerline{\includegraphics[width=1\columnwidth]{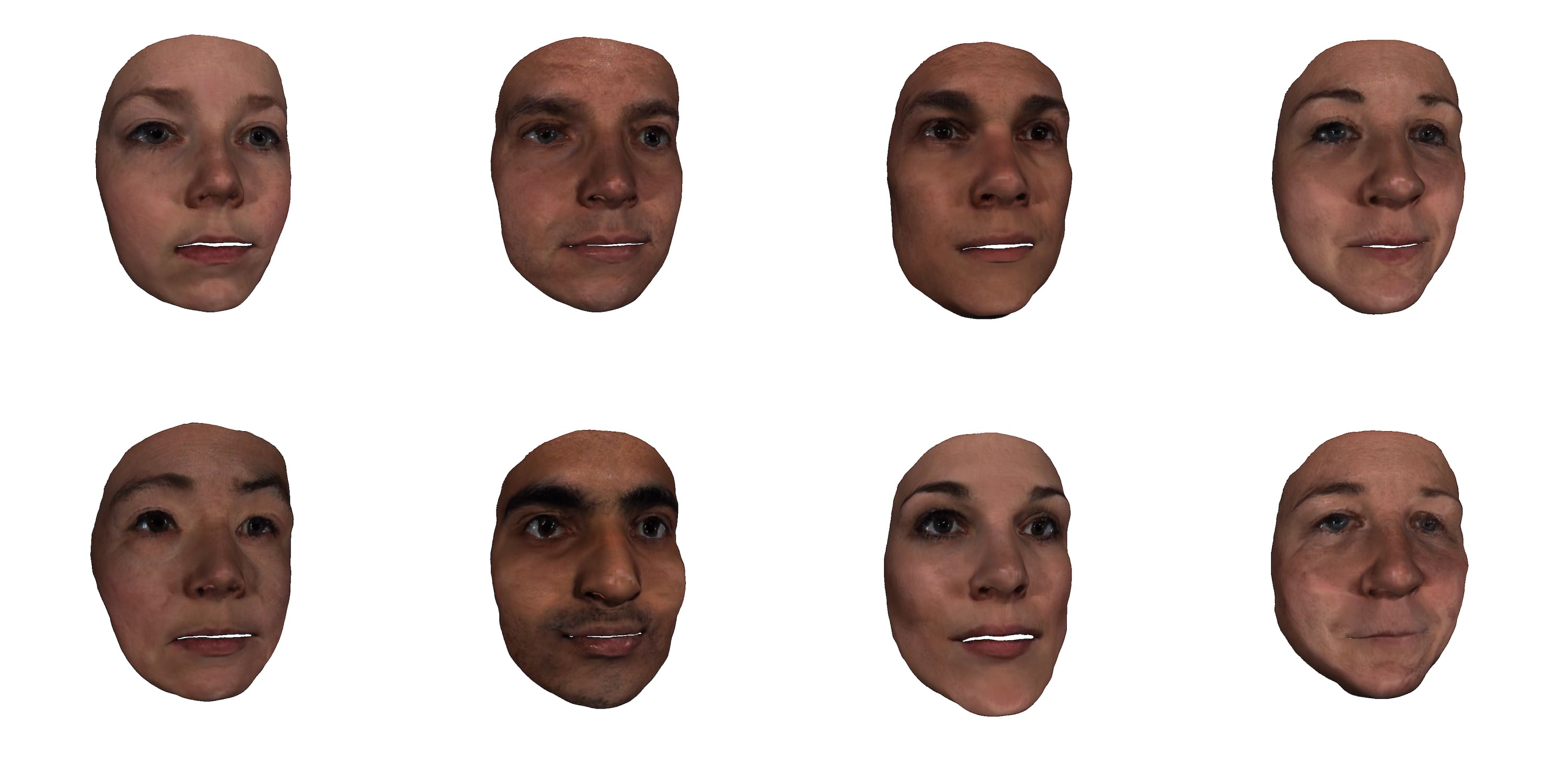}}
\caption{Manipulation results of our method with various inputs and text prompts: `Asian', `Indian', `Woman', `Man'.  The top row shows the original outputs, the bottom row shows the manipulation results, target text prompt is above each column.}
\label{fig:complex}
\end{center}
\vskip -0.3in
\end{figure}

\section{Experiments}
\label{sec:experiments}
In this section, we present the experimental results of our proposed method and evaluate our results based on manipulation quality and fidelity. In addition, we present a simple baseline method for manipulation of 3D objects using PCA, and show that our method allows for more disentangled manipulations without changing the identity of the manipulated face.  Furthermore, we compare the manipulation performance of our method and TBGAN on a list of face expressions. 

\begin{figure*}[!t]
\vskip 0.4in
\hspace{1.7cm} Happy \hspace{1.8cm} Sad \hspace{1.7cm} Surprised  \hspace{1.5cm} Angry  \hspace{1.6cm} Afraid  \hspace{1.5cm} Disgusted
\vspace{-1pt}
\begin{center}
    \begin{tabular}{C{7pt} L{2\columnwidth}}
    \rotatebox[origin=lc]{90}{\centering \hspace{0.7cm} Original} &
    \includegraphics[width=2\columnwidth]{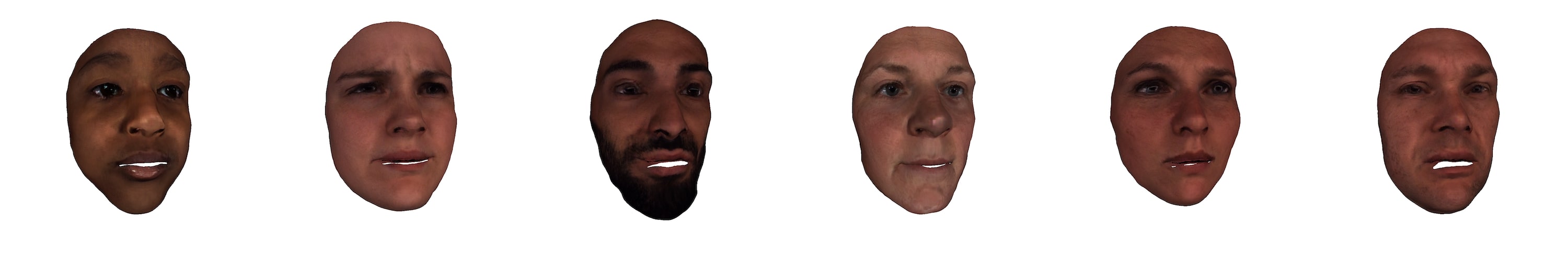} \\
    \rotatebox[origin=lc]{90}{\centering \hspace{0.5cm} TBGAN} &
    \includegraphics[width=2\columnwidth]{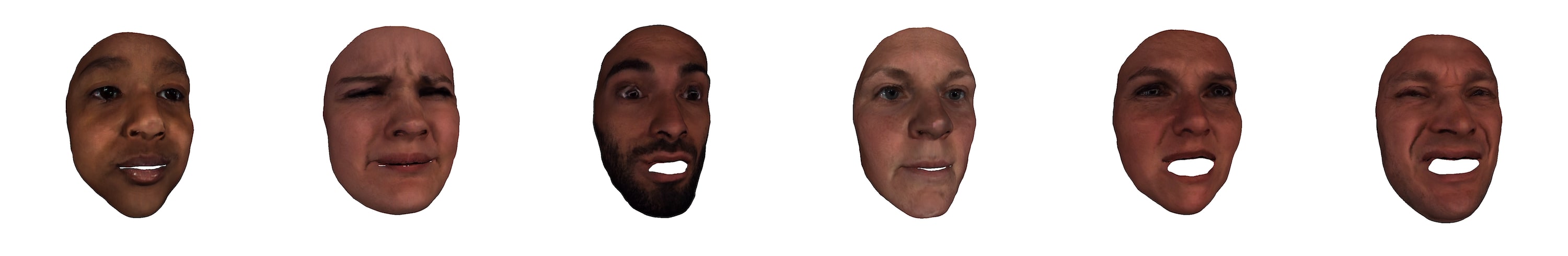} \\
    \rotatebox[origin=lc]{90}{\centering \hspace{0.7cm} Ours} &
    \includegraphics[width=2\columnwidth]{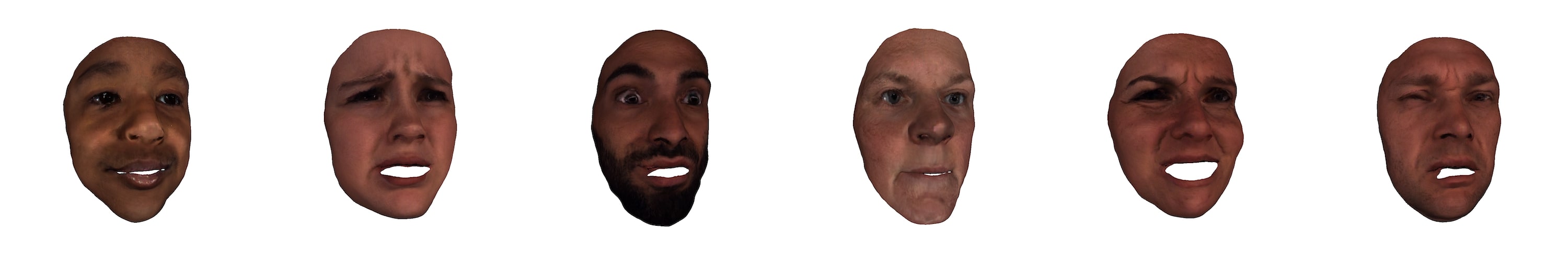} \\
    \end{tabular}
    \caption{The results of our method with text prompts: 'Happy', 'Sad', 'Surprised', 'Angry', 'Afraid', and 'Disgusted'. The top row shows the original outputs, the second row shows the TBGAN conditioned expressions, the third row shows the manipulation results, the target text prompt is above each column.}
    \label{fig:exp}
\vskip -0.3in
\end{center}
\end{figure*}
\subsection{Experimental Setup}
\label{sec:exp-setup}
We use the official implementation and pre-trained model of TBGAN \footnote{\url{https://github.com/barisgecer/TBGAN}}. For all manipulation experiments, we first generate random meshes using TBGAN and its default hyperparameters. For differentiable rendering, we use the renderer of PyTorch3D without additional training or fine-tuning and render 3 images at each generation step with mesh y-axis angles set to $-30$, $3$, and $30$, respectively. The rest of the renderer hyperparameters are as follows: We set the blur radius to $0.0$,  faces per pixel to 2.0, and the position of the point lights is set to $(0.0,\ 0.0,\ +3.0)$.  For all manipulation experiments with CLIP, we set the loss terms as follows: $\lambda_{\text{ ID }}=0.01$, $\lambda_{\text{L2}}=0.001$.  We use a fixed number of $100$ optimization steps for each manipulation. We also use the Adam optimizer and keep the default hyperparameters. We run all experiments on a TITAN RTX GPU.

\subsection{Qualitative Results}
In this section, we demonstrate the quality and consistency of the results obtained by our method on a diverse set of generated faces. We start with simple manipulations such as \textit{`big eyes'} and continue with complex text prompts such as \textit{`Asian'}. We then continue with manipulations on facial expressions and then share the qualitative results for the image-based manipulations.

\paragraph{\textbf{Results on Simple and Complex Text Prompts}}
We begin with a series of text-based manipulations, ranging from simple attributes such as \textit{`big eyes', `thick eyebrows', `makeup'} to fine-grained attributes such as \textit{`old', `child'}, and present the results in Figure \ref{fig:simple}. As can be seen in the figure, our method successfully performs the targeted manipulations on various faces and produces details with high granularity while preserving the global semantics and underlying content. For example, the manipulated outputs for the target text prompt \textit{`old'} represent elderly people, while preserving the overall shape and identity of the original meshes. We also show that our method provides global semantic understanding of more complex attributes such as \textit{`man', `woman', `Asian'}, and \textit{`Indian'}. Figure \ref{fig:complex} shows the results for manipulations on various randomly generated outputs, where we can see that our method is able to perform complex edits such as \textit{ethnicity} and \textit{gender}.

\begin{figure}
{\hspace{0.4cm} $\alpha = 0$ \hspace{1.0cm} $\alpha = +$ \hspace{0.7cm} $\alpha = ++$  \hspace{0.6cm} $\alpha = +++$}
\begin{center}
\includegraphics[width=\columnwidth]{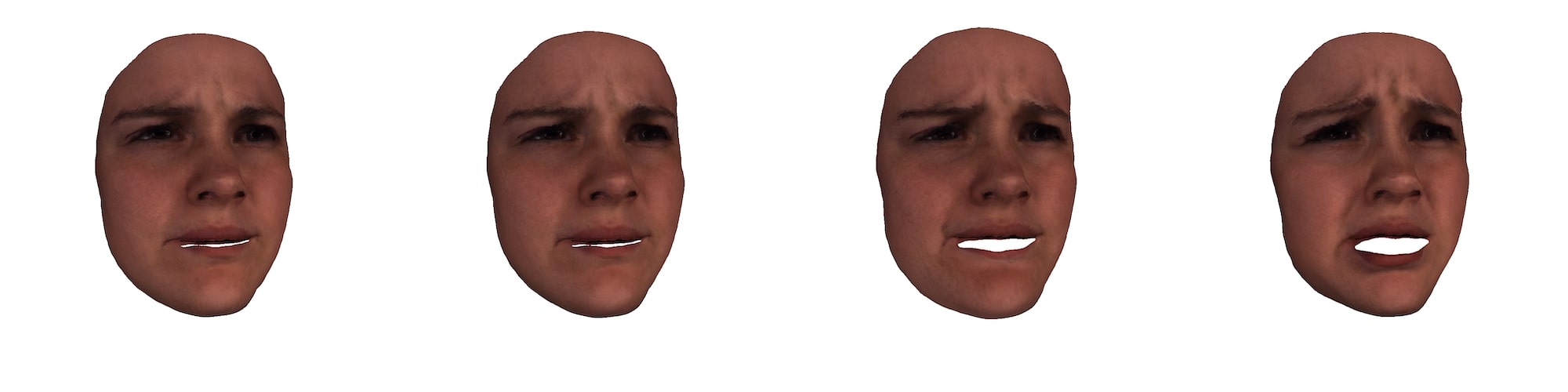}\\
\caption{The results for different manipulation strengths for the text prompt \textit{`sad'}. $\alpha=0$ represents the original image, while $\alpha+$ to $\alpha+++$ represent increasing manipulation strength.}
\label{fig:exp_id}
\vskip -0.1in
\end{center}
\end{figure}

\paragraph{\textbf{Results on Facial Expressions}}
The manipulation capabilities of our method are not limited to the physical characteristics of the generated avatars, but can also be used to change their facial expressions such as \textit{'smiling', 'angry'}, and \textit{'surprised'}. As can be seen in Figure \ref{fig:exp}, our method can successfully manipulate a variety of complex emotions on various input meshes with almost no change to other attributes. Moreover, our model is able to control the intensity of the expression by increasing $\alpha$ such that $\mathbf{c} + \alpha \Delta \mathbf{c}$. As $\alpha$ is increased, the extent of expression changes, as you can see in Figure \ref{fig:exp_id}.

\begin{figure}
\small{\hspace{0.3cm} Original \hspace{0.7cm} Angelina J.  \hspace{0.5cm} C. Ronaldo \hspace{0.5cm} Bill Clinton}
\begin{center}
\includegraphics[width=\columnwidth]{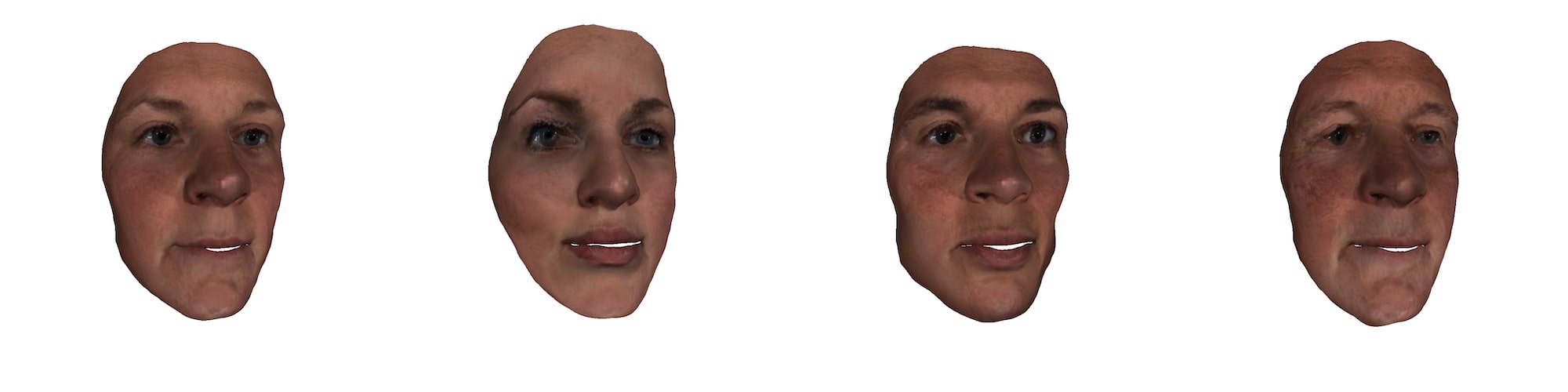}
\caption{Results for text-prompts `Angelina Jolie', `Cristiano Ronaldo' and `Bill Clinton' using our method. }
\label{fig:celeb}
\end{center}
\vskip -0.1in
\end{figure}

\paragraph{\textbf{Results on Identity Manipulations}}
In addition to general manipulations of physical attributes and expressions, we demonstrate that our method can be used to perform complex identity manipulations with solely text-based manipulations. For this experiment, we use text manipulations with the prompts `Bill Clinton', `Cristiano Ronaldo', and `Angelina Jolie' and show the results in Figure \ref{fig:celeb}. As can be seen in the figure, our method is able to achieve a manipulation that captures the characteristics of the target person, such as Bill Clinton's characteristic droopy eyes or Ronaldo's jaw structure.

\begin{figure}
\scriptsize{\hspace{0.5cm} Target \hspace{0.7cm} Original  \hspace{0.6cm} $\lambda_{\text{ID}}=3.0$ \hspace{0.3cm} $\lambda_{\text{ID}}=0.5$ \hspace{0.5cm} $\lambda_{\text{ID}}=0.0$}
\begin{center}
\includegraphics[width=\columnwidth]{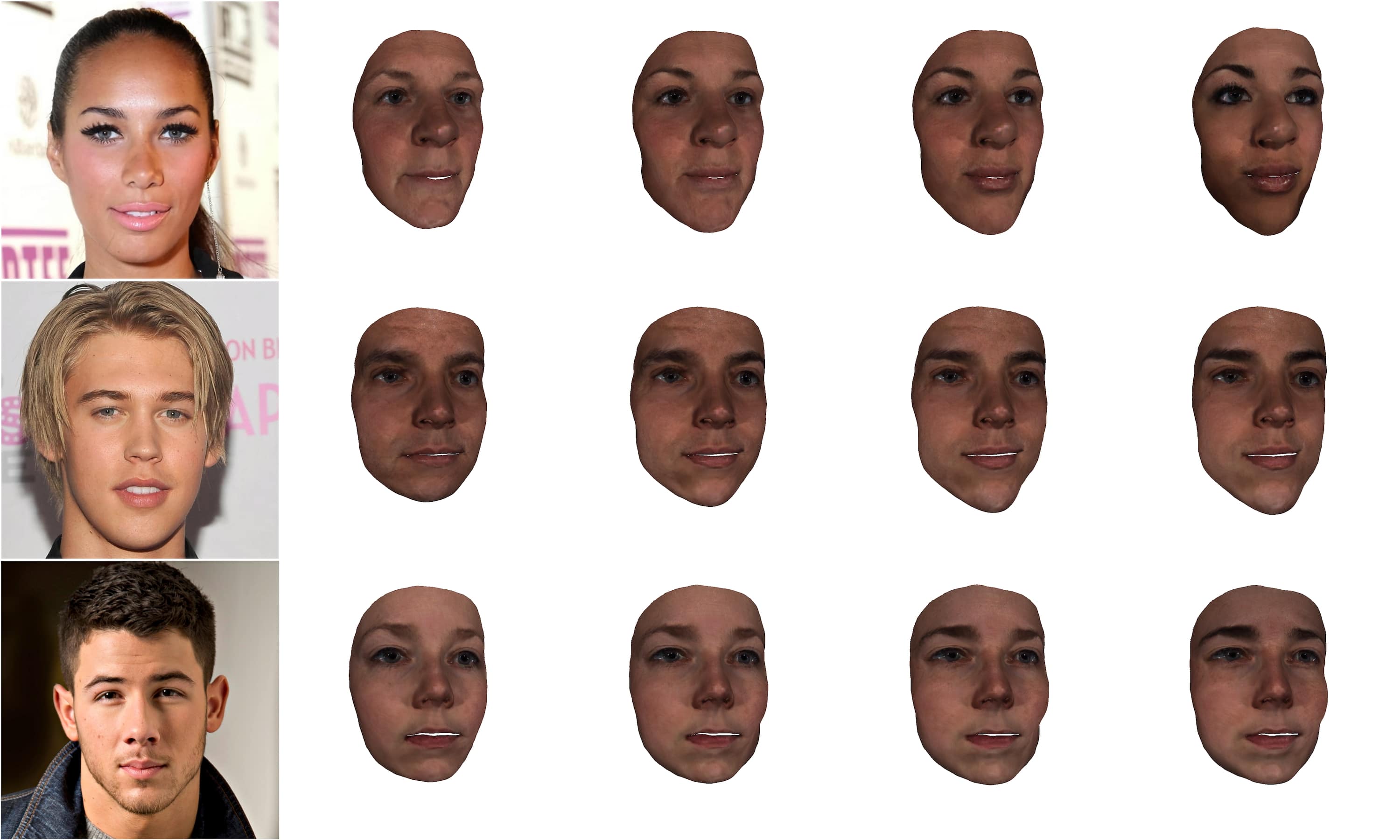}
\caption{Image-based manipulations with the target image on the left and manipulations with different strengths for identity loss on the right.}
\label{fig:image-guided}
\end{center}
\vskip -0.1in
\end{figure}

\paragraph{\textbf{Results on Image-Guided Manipulations}}
We note that our method is not limited to text-based manipulation, but can also be used to manipulate a mesh with a target image, as described in Section \ref{sec:manipulation}. We note that unlike text-based manipulations, image-based manipulations inevitably alter both coarse and fine-grained attributes. Figure \ref{fig:image-guided} shows the manipulation results for different images of celebrities with $\lambda_{\text{L2}}=0$ and $\lambda_{\text{ ID }}=\{3.0, 0.5, 0\}$. As can be seen in the figure, our method can capture complex identity-related attributes of the target image regardless of illumination and pose, and perform successful manipulations such as capturing \textit{skin color, eye makeup, eyebrow} and \textit{jaw structure}. We can also observe that the manipulations with $\lambda_{\text{ ID }}=3.0$ produce a face that resembles both the original image and the target, while $\lambda_{\text{ ID }}=0.0$ reconstructs the target image in 3D.

\begin{figure}[!t]
\centering
\hspace{0.5cm} Original \hspace{0.3cm} Man  \hspace{0.5cm} Woman \hspace{0.5cm} Old \hspace{0.6cm} Child
\begin{center}
    \begin{tabular}{C{7pt} L{2\columnwidth}}
    \rotatebox[origin=lc]{90}{\centering \hspace{0.2cm} PCA} &
    \includegraphics[width=0.9\columnwidth]{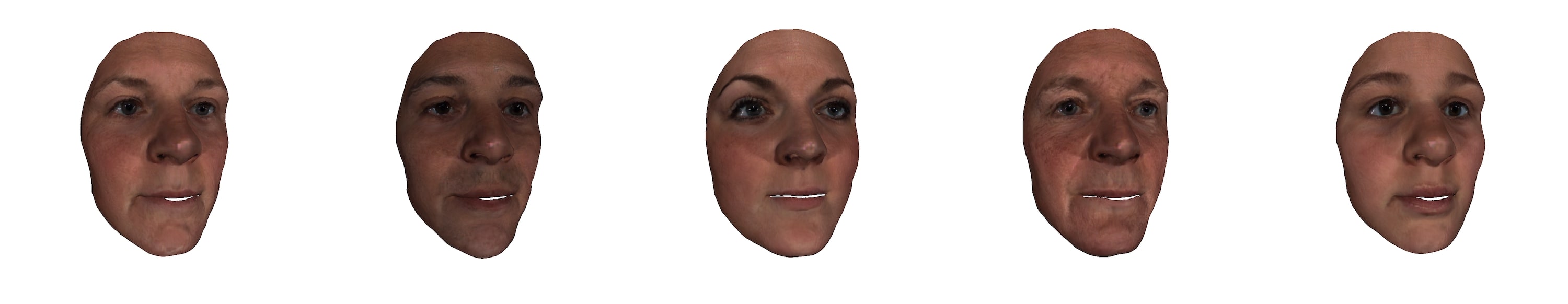} \\
    \rotatebox[origin=lc]{90}{\centering\hspace{0.2cm} Ours} &
    \includegraphics[width=0.9\columnwidth]{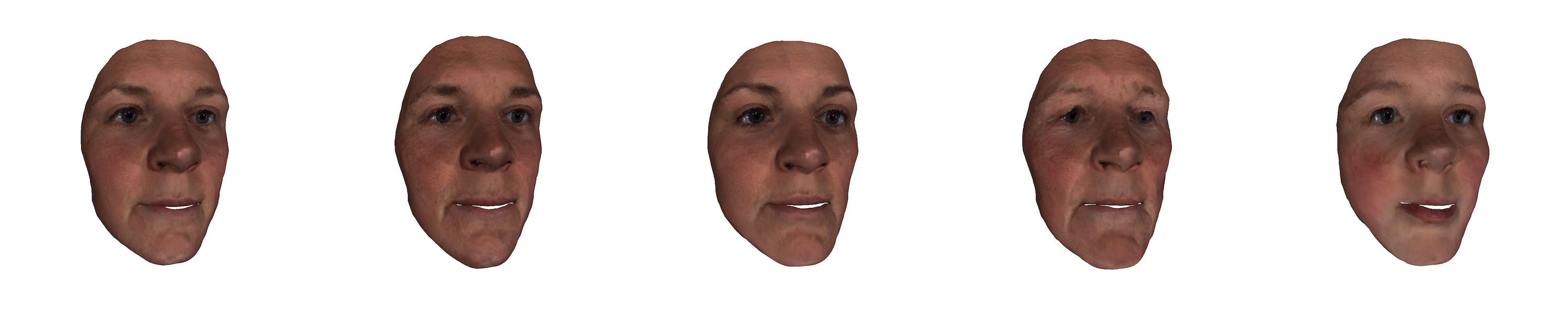} \\
    \end{tabular}
\caption{Comparison between PCA-based manipulations and text-driven manipulations using our method. The top row shows the PCA-based results, the bottom row shows the results with our method.}
\label{fig:pca}
\end{center}
\vskip -0.1in
\end{figure}

\begin{table*} [!t]
\begin{center}
\begin{tabular}{cccccc}
\toprule
 \textbf{Accuracy} &  Male & Female & Old & Young & All \\
\hline PCA & $\mathbf{4.42}$ \footnotesize $\pm 0.65$ & $\mathbf{4.13}$ \footnotesize $\pm 0.99$ & $4.25$ \footnotesize $\pm 0.61$ & $4.21$ \footnotesize $\pm 0.83$ & $4.25$ \footnotesize $\pm 0.78$ \\
Ours & $4.21$ \footnotesize $\pm 0.83$ & $4.08$ \footnotesize $\pm 0.97$ & $\mathbf{4.67}$ \footnotesize $\pm 0.56$ & $\mathbf{4.46}$ \footnotesize $\pm 0.78$ & $\mathbf{4.35}$ \footnotesize $\pm 0.82$ \\
\toprule \textbf{Identity} &  Male & Female & Old & Young & All \\\hline
PCA & $2.85$ \footnotesize $\pm 1.04$  & $2.57$ \footnotesize {$\pm 1.08$} & $3.52$ \footnotesize $\pm 1.29$ & $3.01$ \footnotesize $\pm 1.30$ & $2.99$ \footnotesize $\pm 1.21$ \\
Ours & $\mathbf{4.45}$ \footnotesize $\pm 0.76$ & $\mathbf{4.33}$ \footnotesize $\pm 0.80$ & $\mathbf{3.70}$ \footnotesize $\pm 1.22$ & $\mathbf{3.62}$ \footnotesize$\pm 1.20$ & $\mathbf{4.02}$ \footnotesize $\pm 1.07$ \\
\bottomrule
\end{tabular}
\caption{Mean scores (1-5) for identity preservation and manipulation accuracy for our method and PCA-based baseline.}
\label{tab:human-eval}
\end{center}
\end{table*}

\subsection{Comparison with PCA}
Our method performs a text-driven manipulation of facial meshes. Since there are no existing approaches for this task, we propose a simple PCA-based baseline inspired by GANSpace \cite{harkonen2020ganspace}.  For this experiment, we sample $10,000$ intermediate latent vectors from the $4\times4$/dense layer of TBGAN and apply PCA to the concatenated latent vectors to obtain principal components, where each component represents a new transformed axis that is a linear combination of the original features. Using the found directions, we can manipulate the original latent vectors by directly applying a principal component, i.e., using the component as a direction vector to steer the generation:

\begin{equation} 
    c^{'} = c + (\alpha \times n \times \text{PC}_{i})            
\end{equation}

Here $\alpha$ denotes the step size, $n$ the number of steps, and $\text{PC}_{i}$ the $i_{\text{th}}$ principal component used. For comparison purposes, we keep the top ranked principal components and apply to them randomly generated latent vectors with a step size of $\alpha$ of 10. We note that the top-ranking principal components encode prominent directions such as \textit{age} and \textit{gender}. For comparison, we apply age and gender-based text-driven edits to the same latent vectors and present the comparative results in Figure \ref{fig:pca}. As can be seen in the figure, the top directions encoded by PCA, \textit{age} and \textit{gender}, significantly alter the identity of the input person, while our method achieves the desired manipulations without altering irrelevant attributes.

\vspace{-0.3cm}
\paragraph{\textbf{Human Evaluation}}
We also conduct human evaluations to measure the perceived quality of the generated results. More specifically, we are interested in the extent to which the manipulation results match the input target text and preserve the other attributes. For the human evaluation, we asked $n=25$ users to evaluate 5 randomly selected sets of input meshes, text prompts $t$, and manipulated meshes. For each set, we display the target text and output in pairs and ask users to assign a score between $[1, 5]$ for two questions: \emph{`How well does the manipulation achieve the attributes specified in the target text?'} and \emph{`Is the identity of the input face preserved?'}. In Table \ref{tab:human-eval}, we report the mean and standard deviations of the scores for the questions (denoted as  \textit{Accuracy} and \textit{Identity}, respectively). As the human-evaluation results show, our method performs better than PCA in all settings except the Female and Male attributes. However, we note that our method performs significantly better than PCA on these attributes when it comes to identity preservation, which was evaluated by human raters. More specifically, the human raters found that our method preserves identity while achieving competitive attribute preservation scores that are $34\%$ higher than PCA.

\begin{table} [!t]
\begin{center}
\begin{tabular}{ccc}
\toprule & TBGAN & Ours \\
\hline Surprise & $4.64$ \footnotesize$\pm 0.48$ & $\mathbf{4.70}$ \footnotesize$\pm 0.46$ \\
\hline Disgust & $4.04$ \footnotesize$\pm 1.12$ & $\mathbf{4.26}$ \footnotesize$\pm 0,61$ \\
\hline Happy & $1.83$ \footnotesize$\pm 0.92$ & $\mathbf{3.77}$ \footnotesize$\pm 0.90$ \\
\hline Sad & $4.17$ \footnotesize$\pm 0.87$ & $\mathbf{4.48}$ \footnotesize$\pm 0.71$ \\
\hline Afraid & $2.87$ \footnotesize$\pm 1.08$ & $\mathbf{3.43}$ \footnotesize$\pm 0.97$ \\
\hline Angry & $2.05$ \footnotesize$\pm 1.22$ & $\mathbf{3.65}$ \footnotesize$\pm 1.24$ \\
\hline All & $3.26$ \footnotesize$\pm 1.46$ & $\mathbf{4.05}$ \footnotesize$\pm 0.97$ \\
\bottomrule
\end{tabular}
\caption{Mean scores (1-5) and std values for manipulation accuracy on various expressions using our method and TBGAN.}
\vspace{-0.6cm}
\label{tab:human-eval-exp}
\end{center}
\end{table}

\subsection{Comparison with TBGAN}
TBGAN does not provide a way to manipulate the generated meshes, but it is possible to obtain different facial expressions by modifying  one-hot encoded expression vector of TBGAN. Therefore, we compare the manipulations done by our method on the facial expressions with TBGAN. As can be seen in Figure \ref{fig:exp}, our results are more successful in terms of realistic facial expression representation. We also performed a human evaluation to compare the performance of facial expression manipulations by asking $n=25$ participants; \emph{"How well does the manipulation achieve the expressions specified in the target text?"}. Table \ref{tab:human-eval-exp} shows the mean scores and standard deviations of the results, with our method outperforming TBGAN in all settings. Moreover, our method is able to gradually change the expression (see Figure \ref{fig:exp_id}) which is not possible using TBGAN since it only produces a fixed expression using the one-hot encoded vector.

\section{Limitations and Broader Impact}
\label{sec:limitations}
While our method is very effective for both coarse and fine-grained manipulations, our base generative model is trained to generate partial face avatars. We therefore strongly believe that our work can be extended to more comprehensive generation scenarios to produce full head meshes or bodies.

\section{Conclusion}
\label{sec:conclusion}
We proposed an approach to manipulating 3D facial avatars with text and image inputs that relies on a differentiable rendering pipeline combined with CLIP -based and identity-based losses. Unlike previous work that limits the manipulation of 3D shapes to either local geometric changes such as texture or only shape, our method can perform high-level and complex manipulations of shapes and textures. Another major advantage of our method is that it requires only 5 minutes to manipulate a given mesh, while other works require an hour of optimization time per manipulation. Given that avatar and human body generation is widely used in industries such as character design, animation, and visual effects, we see two natural improvements for our work: sketch-based manipulation for more intuitive and user-friendly manipulations, and the extension of our framework to full-body generation.\\

\noindent \textbf{Acknowledgments}
This research is produced with support from the 2232 International Fellowship for Outstanding Researchers Program of TUBITAK (Project No: 118c321).

{\small
\bibliographystyle{ieee_fullname}
\bibliography{arxiv}
}

\appendix

\section{Ablation Study}
In this section, we perform ablation studies on the effects of identity loss and layer selection for latent space manipulation. 

\subsection{Effect of Layer Selection}
\label{app:layer}
We perform our manipulations on the $4\times4$/Dense layer of TBGAN, the layer that provides the best results in terms of identity preservation and meaningful manipulations. 

\begin{figure}[!t]
\small{\hspace{0.5cm} Original \hspace{0.2cm} 4x4/Dense  \hspace{0.5cm} 8x8 \hspace{0.7cm} 32x32 \hspace{0.4cm} 128x128}
\centering
\begin{center}
    \begin{tabular}{C{7pt} L{2\columnwidth}}
    \rotatebox[origin=lc]{90}{\centering \hspace{0.4cm}Old} &
    \includegraphics[width=0.9\columnwidth]{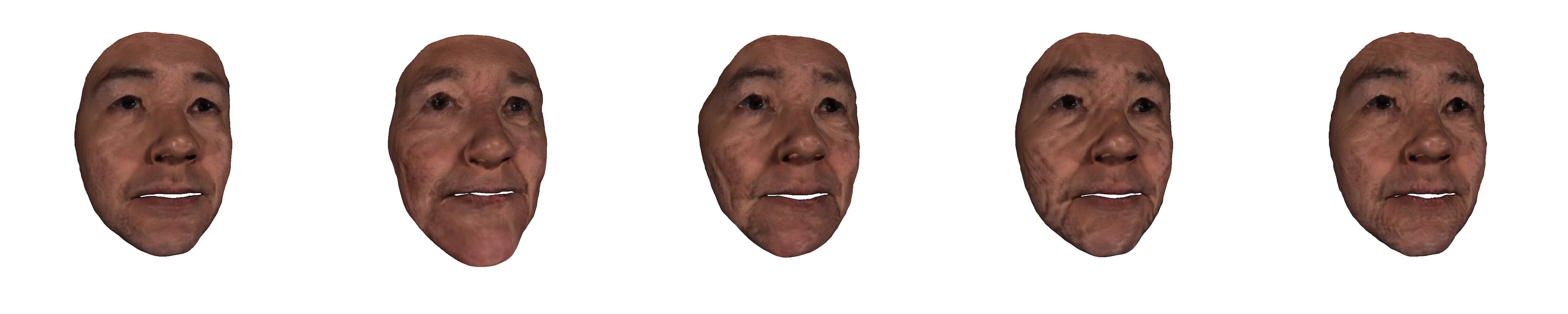} \\
    \rotatebox[origin=lc]{90}{\centering\hspace{0.3cm}Beard} &
    \includegraphics[width=0.9\columnwidth]{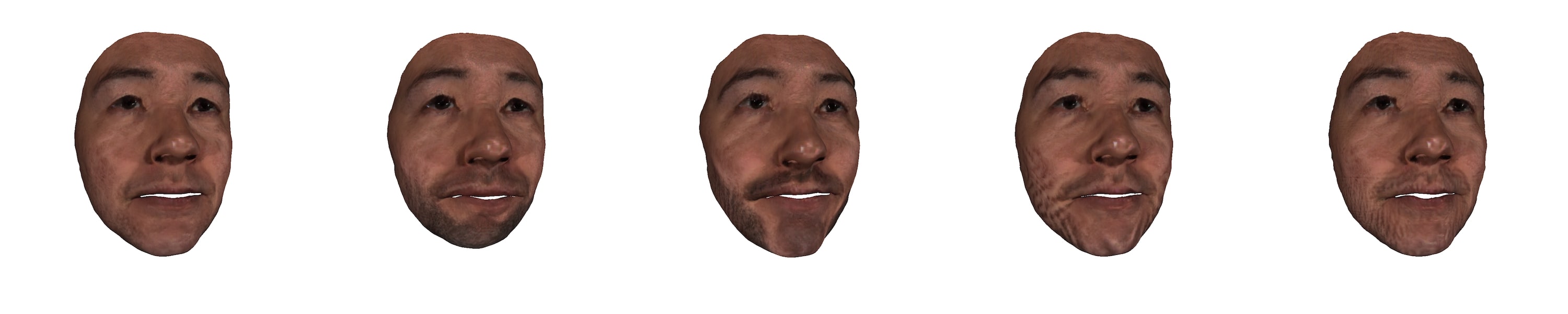} \\
    \rotatebox[origin=lc]{90}{\centering\hspace{0.4cm}Old} &
    \includegraphics[width=0.9\columnwidth]{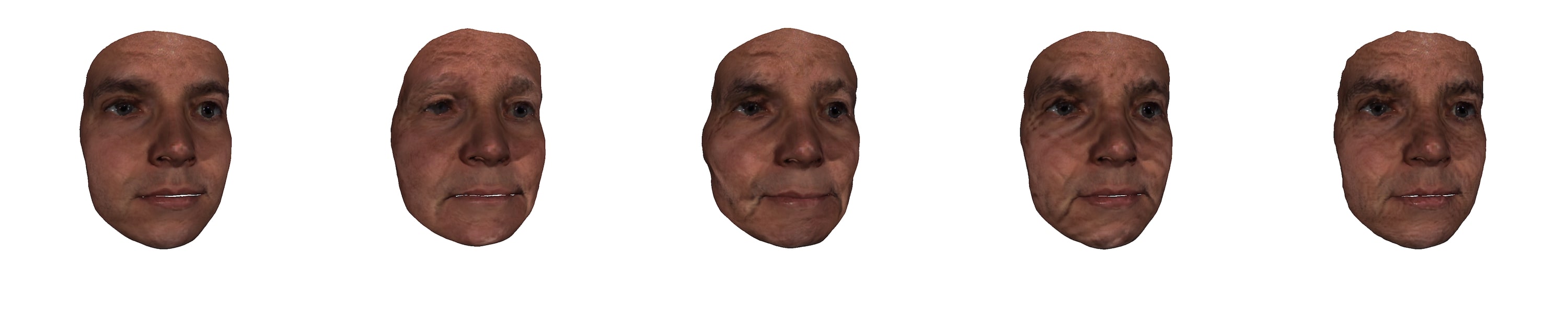} \\
    \rotatebox[origin=lc]{90}{\centering\hspace{0.3cm}Beard} &
    \includegraphics[width=0.9\columnwidth]{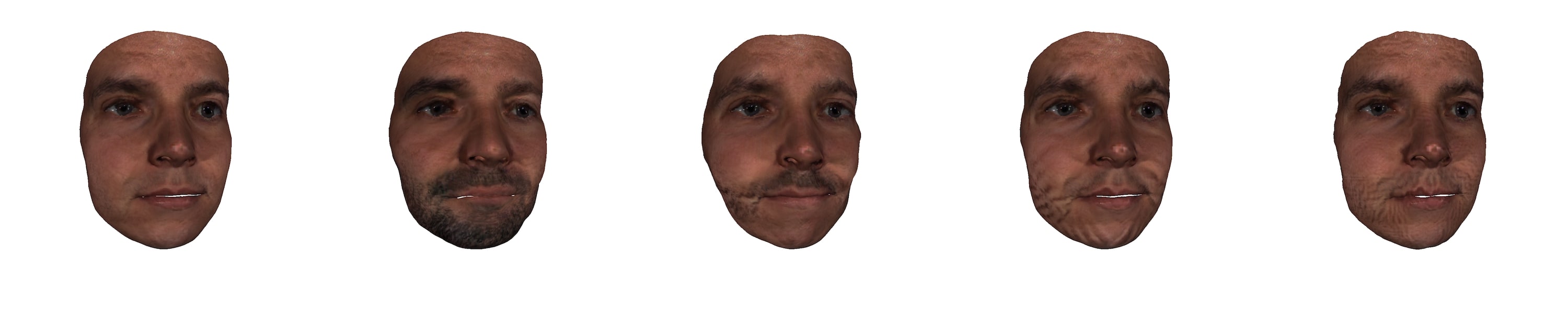} \\
\end{tabular}
\caption{The comparison of manipulations on different layers for two different 3D faces. First two column show the `beard' and `old' manipulations on one 3D face and the second column show the results for the same manipulations on another 3D face. }
\label{fig:layers}
\end{center}
\vskip -0.1in
\end{figure}

The comparison of our method on different layers can be found in Figure \ref{fig:layers}. We show that the manipulations on other layers give defected results with undesirable artifacts, so that the results deviate from the desired text prompt.

\subsection{Effect of Identity Loss}
\label{app:identity}

\begin{figure}[ht]
\vskip 0.1in
\begin{minipage}[b]{0.45\linewidth}
    \centering
        \begin{tabular}{C{7pt} L{2\columnwidth}}
            \rotatebox[origin=lc]{90}{\centering\tiny{\hspace{0.2cm} Original}} &
            \includegraphics[width=\textwidth]{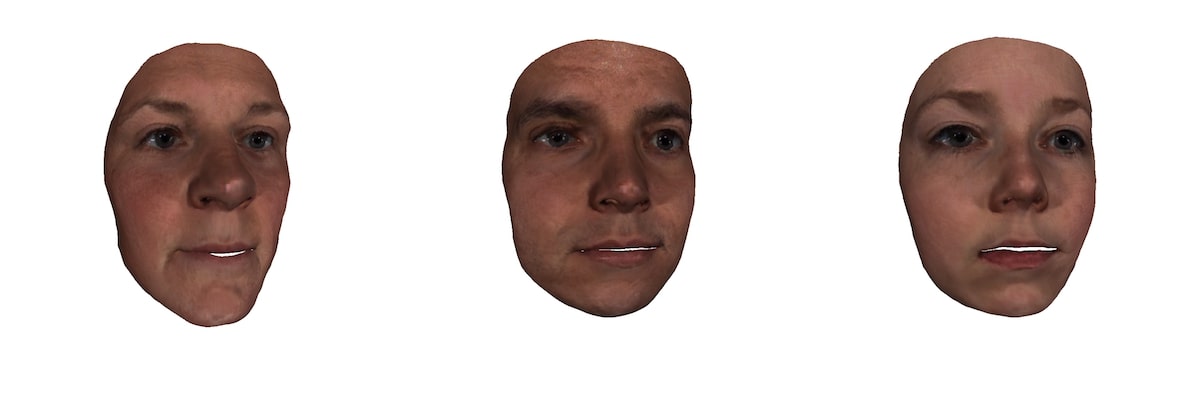} \\
            \rotatebox[origin=lc]{90}{\centering\tiny{\hspace{0.3cm}Child}} &
            \includegraphics[width=\textwidth]{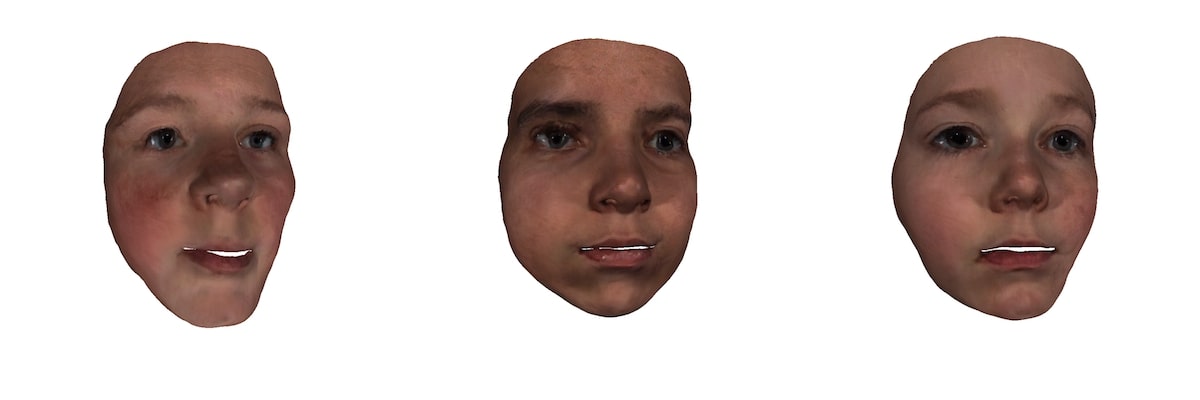} \\
            \rotatebox[origin=lc]{90}{\centering\tiny{\hspace{0.3cm}Makeup}} &
            \includegraphics[width=\textwidth]{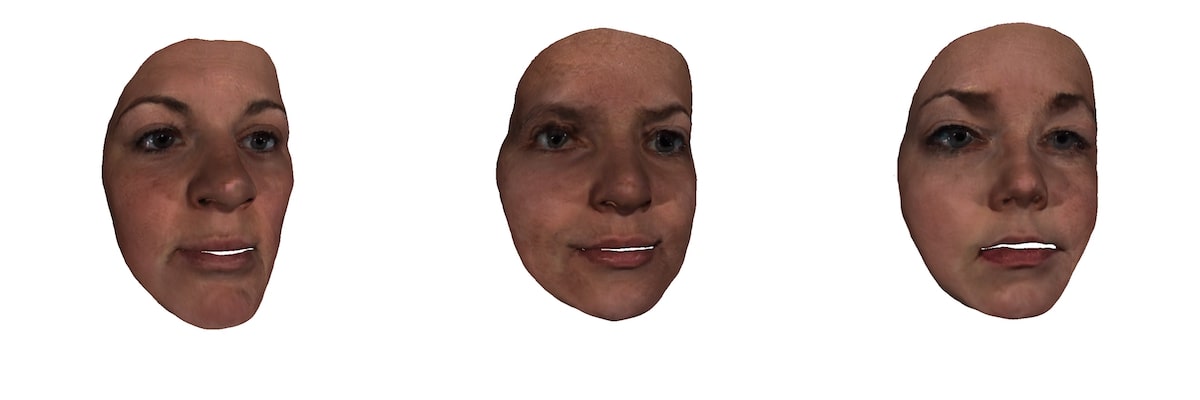} \\
            \rotatebox[origin=lc]{90}{\centering\tiny{\hspace{0.2cm}Big Eyes}} &
            \includegraphics[width=\textwidth]{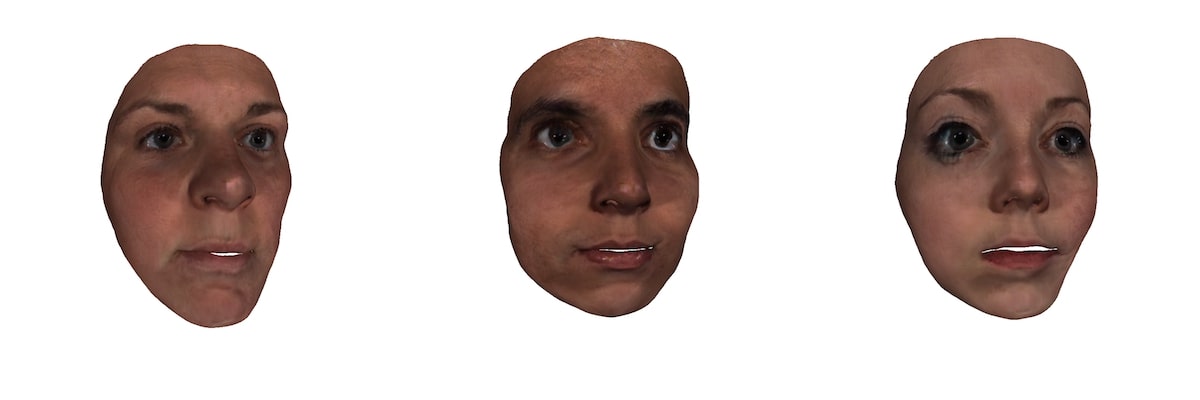} \\
        \end{tabular}
    \centering\footnotesize{(a) With ID loss. }
    \label{fig:with_id}
\end{minipage}
\hspace{0.5cm}
\begin{minipage}[b]{0.45\linewidth}
    \centering
        \begin{tabular}{C{7pt} L{2\columnwidth}}
            \rotatebox[origin=lc]{90}{} &
            \includegraphics[width=\textwidth]{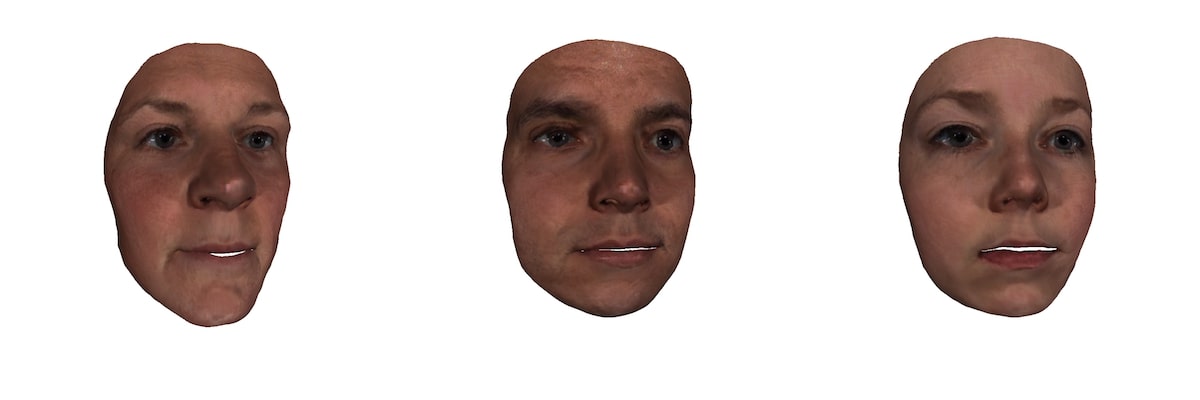} \\
            \rotatebox[origin=lc]{90}{} &
            \includegraphics[width=\textwidth]{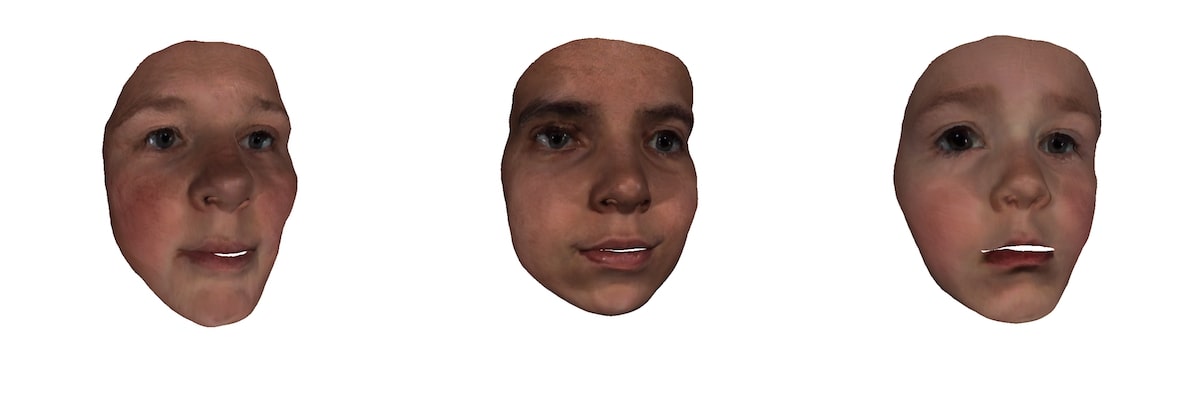} \\
            \rotatebox[origin=lc]{90}{} &
            \includegraphics[width=\textwidth]{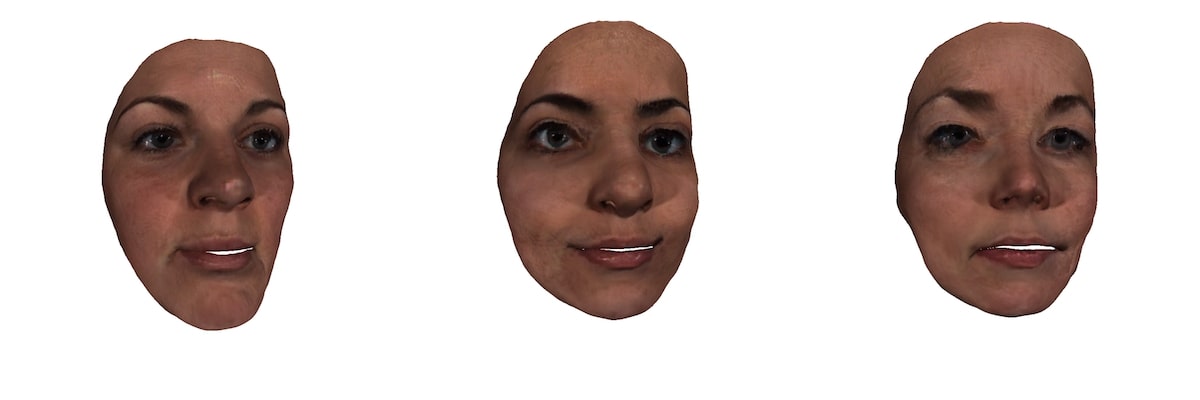} \\
            \rotatebox[origin=lc]{90}{} &
            \includegraphics[width=\textwidth]{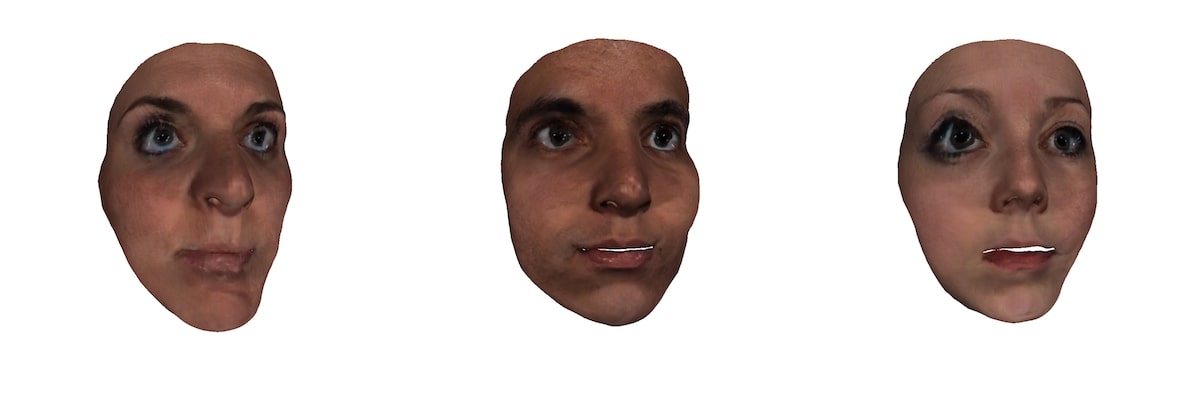} \\
        \end{tabular}
    \centering\footnotesize{(b) Without ID loss. }
    \label{fig:without_id}
\end{minipage}
\caption{Results with and without ID loss. }
\label{fig:id_loss}
\vskip -0.3in
\end{figure}

Our method uses ArcFace, a large-scale pre-trained face recognition network, to compute identity loss $\mathcal{L}_{\text{ ID }}$ and enforce identity preservation during manipulation. We perform an ablation study with different target texts describing emotion-, shape-, and texture-related changes to demonstrate the effect of $\mathcal{L}_{\text{ ID }}$ on the manipulation results, and present the results in Figure \ref{fig:id_loss}. For the identity loss experiments, we simply set $\mathcal{L}_{\text{ ID }}=0$ and leave the other hyperparameters the same. As can be seen in Figure \ref{fig:id_loss}, identity loss is crucial for preserving the identity of the input, and omitting it leads to manipulation results that are significantly different from the input.

\begin{table} [!t]
\begin{center}
\begin{tabular}{cc}
\hline `a bad photo of a' & `a sculpture of a' \\
\hline `a photo of the hard to see' &  `a low resolution photo of the' \\ 
\hline `a rendering of a' & `graffiti of a' \\
\hline `a bad photo of the' &  `a cropped photo of the' \\
\hline `a photo of a hard to see' & `a bright photo of a' \\
\hline `a photo of a clean' & `a photo of a dirty' \\
\hline `a dark photo of the' & `a drawing of a' \\ 
\hline `a photo of my' & `the plastic' \\ 
\hline `a photo of the cool' &  `a close-up photo of a' \\
\hline `a painting of the' & `a painting of a' \\
\hline `a pixelated photo of the' & `a sculpture of the' \\
\hline `a bright photo of the' & `a cropped photo of a' \\ 
\hline `a plastic' &  `a photo of the dirty' \\
\hline `a blurry photo of the' & `a photo of the' \\
\hline `a good photo of the' & `a rendering of the' \\
\hline `a in a video game.' & `a photo of one' \\
\hline `a doodle of a' &  `a close-up photo of the' \\
\hline `a photo of a' &  `the in a video game.' \\
\hline `a sketch of a' &  `a face of the' \\
\hline `a doodle of the' & `a low resolution photo of a' \\
\hline `the toy' & `a rendition of the' \\
\hline `a photo of the clean' &`a photo of a large' \\
\hline `a rendition of a' & `a photo of a nice' \\
\hline `a photo of a weird' & `a blurry photo of a' \\
\hline `a cartoon' &`art of a' \\
\hline `a sketch of the' & `a pixelated photo of a' \\
\hline `itap of the' & `a good photo of a' \\
\hline `a plushie' & `a photo of the nice' \\
\hline `a photo of the small' &`a photo of the weird' \\
\hline 'the cartoon' &`art of the' \\
\hline `a drawing of the' & `a photo of the large' \\
\hline `the plushie' & `a dark photo of a' \\
\hline `itap of a' & `graffiti of the' \\
\hline `a toy' & `itap of my' \\
\hline `a photo of a cool' & `a photo of a small' \\
\hline `a 3d object of the' &`a 3d object of a' \\
\hline `a 3d face of a' & `a 3d face of the' \\
\hline \end{tabular}
\caption{List of templates that our method uses for augmentation. The input text prompt is added to the end of each sentence template.}
\label{tab:templates-prompt}
\end{center}
\end{table}

\section{Sentence Templates for Prompt Engineering}
\label{app:templates}

Our method uses 74 sentence templates. The list of templates we use for augmentation can be found in Table \ref{tab:templates-prompt}.

\end{document}